\documentclass[10pt,twocolumn,letterpaper]{article}

\usepackage{iccv}
\usepackage{times}
\usepackage{epsfig}
\usepackage{graphicx}
\usepackage{amsmath}
\usepackage{amssymb}
\usepackage{etoolbox}

\iccvfinalcopy 

\def\iccvPaperID{****} 

\usepackage[utf8]{inputenc} 
\usepackage[T1]{fontenc}    
\usepackage{url}            
\usepackage{booktabs}       
\usepackage{amsfonts}       
\usepackage{nicefrac}       
\usepackage{microtype}      
\usepackage{float}
\usepackage{caption}
\usepackage{amsmath}
\usepackage[square,numbers]{natbib}
\usepackage[pagebackref=true,breaklinks=true,letterpaper=true,colorlinks,bookmarks=false]{hyperref}
\usepackage{xcolor}
\usepackage{footnote}

\usepackage{makecell}
\AtBeginEnvironment{tabular}{\footnotesize}

\title{SqueezeNAS: Fast neural architecture search for faster semantic segmentation\vspace{-0.1in}}
\author{\large Albert Shaw, Daniel Hunter, Forrest Iandola and Sammy Sidhu \vspace{0.05in} \\
\vspace{0.05in}
\large DeepScale Inc.\\
{\tt\small\{albert,daniel,forrest,sammy\}@deepscale.ai}
}

\begin{document}

\setlength{\abovedisplayskip}{5pt} 
\setlength{\belowdisplayskip}{5pt} 

\maketitle

\begin{abstract}
\vspace{-0.1in}

For real time applications utilizing Deep Neural Networks (DNNs), it is critical that the models achieve high-accuracy on the target task and low-latency inference on the target computing platform.
While Neural Architecture Search (NAS) has been effectively used to develop low-latency networks for image classification, there has been relatively little effort to use NAS to optimize DNN architectures for other vision tasks. 
In this work, we present what we believe to be the first proxyless hardware-aware search targeted for dense semantic segmentation.
With this approach, we advance the state-of-the-art accuracy for latency-optimized networks on the Cityscapes semantic segmentation dataset.
Our latency-optimized small SqueezeNAS network achieves 68.02\% validation class mIOU with less than 35 ms inference times on the NVIDIA Xavier. 
Our latency-optimized large SqueezeNAS network achieves 73.62\% class mIOU with less than 100 ms inference times. 
We demonstrate that significant performance gains are possible by utilizing NAS to find networks optimized for both the specific task and inference hardware. 
We also present detailed analysis comparing our networks to recent state-of-the-art architectures.

\end{abstract}
\vspace{-0.2in}

\section{Introduction and Motivation}
\label{sec:introduction_and_motivation}
\vspace{-0.08in} 

In recent years, Deep Neural Networks (DNNs) have become a dominant approach for solving numerous problems in computer vision. 
Image classification tasks such as ImageNet~\cite{Imagenet} and CIFAR10~\cite{CIFAR} are the de facto "playground" for designing DNN model architectures.
When developing DNNs for a target task other than image classification (e.g. semantic segmentation or object detection), a popular approach is to use {\em architecture-transfer}:
start with an image classification network and append a few task-specific layers to the end of the network. \footnote{In our terminology, we refer to the task-specific end of the network as the {\em head}, and we refer to the portion of the network that was originally designed for image classification as the {\em backbone}.}

We believe architecture-transfer has become mainstream because of a number of conventional-wisdom assumptions that have permeated the computer vision community.
In the following, we enumerate these assumptions and present evidence for why these assumptions are becoming outdated.

\begin{itemize}
    \vspace{-3mm}
	\item {\bf Assumption 1: The most accurate neural network for ImageNet image classification will also be the most accurate backbone for the target task.}\\ 
	Reality: ImageNet accuracy is only loosely correlated with accuracy on a target task. 
	For example, SqueezeNet is a small neural network that achieves significantly lower ImageNet classification accuracy than VGG~\cite{squeezenet}~\cite{VGGNet}. 
	However, SqueezeNet is more accurate than VGG when used for the task of identifying similar patches in a set of images~\cite{zhang2018perceptual}. 
	Thus, the right DNN design varies depending on the target task.

	\vspace{-0.1in}
	\item {\bf Assumption 2: Neural Architecture Search (NAS) is prohibitively expensive.} \\
	Reality: It is true that some NAS methods based on genetic algorithms (e.g.~\cite{AmoebaNet}) or reinforcement learning (e.g.~\cite{zoph2016neural}) often require thousands of GPU days to converge on a good DNN design because they train hundreds or thousands of different DNNs before converging. 
	However, recent "supernetwork" approaches such as DARTS~\cite{DARTS} and FBNet~\cite{FBNET} have turned the problem inside out. They can train one supernetwork that contains millions of DNN designs, but it still converges on an optimal DNN design within 10 GPU days.

\end{itemize}
\vspace{-0.1in}
So, the "right" DNN design depends on the target task, and modern NAS methods can quickly converge on the right DNN for a task.
A similar issue arises when we look at choosing the right DNN for a target computing platform (e.g. a specific version of a CPU, GPU, or TPU):

\begin{itemize}
    \vspace{-3mm}
	\item {\bf Assumption 3: Fewer multiply-accumulate (MAC) operations will yield lower latency on a target computing platform.} \\
	Reality: 
	In a recent study, Almeida \etal showed that two DNNs with the same number of MACs can have a 10x difference in latency on the same computing platform~\cite{EMBench}.
	Further, when the FBNet authors optimized networks for different smartphones, they found a DNN that ran fast on the iPhone X, but slow on the Samsung Galaxy S8; as well as a DNN ran fast on the iPhone, but slow on the Samsung~\cite{FBNET}.
	Depending on the processor and the kernel implementations, different convolution dimensions run faster or slower, even when the number of MACs is held constant.
	
\end{itemize}

To make use of these new realities, we propose a playbook for producing the lowest-latency, highest-accuracy DNNs on a target task and a target computing platform:

\vspace{-0.1in}
\begin{enumerate}
	\item Run Neural Architecture Search directly on the target task (e.g. object detection or semantic segmentation), and not on a proxy task (e.g. image classification).\footnote{If you wish to use outside data from an other task for pretraining, first perform a proxyless search to produce the DNN architecture, then reset the weights and do pretraining on outside data, and finally finetune on the target task.}
	
	\vspace{-0.1in}
	\item Use modern supernetwork-based NAS, and enjoy the fact the search converges quickly.
	
	\vspace{-0.1in}
	\item Configure the NAS to optimize for both accuracy (on the target task) and latency (on the target platform).
\end{enumerate}
\vspace{-0.1in}

In the rest of this paper, we investigate the effectiveness of this playbook by doing a proxyless search using the Cityscapes semantic segmentation dataset~\cite{Cityscapes}, targeting low-latency inference on the NVIDIA Xavier embedded GPU computing platform~\cite{NVIDIA_Xavier}, and producing fast and accurate DNNs.
We refer to the optimized DNNs generated in this study as {\em SqueezeNAS} networks.

\section{Related work}
\label{sec:related_work}
\vspace{-0.08in} 

\subsection{Semantic Segmentation}
\vspace{-0.08in} 
Semantic segmentation is the computer vision task of assigning a class for each pixel in a given image. It is a workhorse in many computer vision applications areas, from automotive (segmenting the road and lane lines) to aerial imagery analysis.
To train and evaluate semantic segmentation models, a number of datasets have been developed such as Cityscapes\cite{Cityscapes}, ADE20k\cite{ADE20k}, NYUDv2\cite{NYUDv2}, and PASCAL VOC\cite{PASCAL} which have made the research in semantic segmentation algorithms much more accessible.

DNNs initially found success with image classification tasks; AlexNet\cite{AlexNet} and its successors dramatically increased the state-of-the-art accuracies on the ImageNet and CIFAR10 classification tasks. Following this success, Long et al. developed Fully Convolutional Networks for Semantic Segmentation\cite{FCN} (FCN) by utilizing an Imagenet backbone - achieving then state-of-the-art performance on VOC PASCAL and NYUDv2.
DeepLab\cite{deeplabv1} later leveraged dilated convolutions to further increase the accuracy on segmentation benchmarks.
The typical workflow of these approaches is to start with an image classification DNN and then adapt it for higher resolution, increasing the compute proportionally to the number of pixels.
This part is usually called the encoder or backbone.
The semantic segmentation network's decoder uses the low resolution feature maps from the encoder to perform more computation and generates an output prediction for each pixel that is the same size as the input resolution.
This decoder or "head" can be a series of deconvolutions like in FCN, or something much more complex like the dilated Spatial Pyramid Pooling (ASPP) module seen in the DeepLab\cite{deeplabv1, deeplabv3, DeepLabV3+} Family. 

Semantic segmentation, however, is a very different task from image classification.
One way semantic segmentation networks differs from image classification networks is that they usually requires much higher resolution inputs to get good results. Image classification networks commonly use an input at a $224 x 224$ resolution, while segmentation networks often use more than 40 times the number of pixels. Segmentation networks also typically have exotic architectures due to the fact that they have a dense high resolution output. Large input resolutions also means that segmentation networks often use trillions of Multiply-Accumulates (MACs) for a single image prediction, whereas accurate image classification networks are usually in the tens of billions.
Many early deep learning approaches focused on maximizing accuracy, without a regard to the number of operations or latency.

\subsection{Efficient Network Design}
\vspace{-0.08in} 
From 2012 to 2016, a substantial portion of the computer vision research community focused on designing DNNs that achieved the highest possible accuracy on image classification. These networks were then modified and finetuned to perform other tasks such as object detection and semantic segmentation. This led to significant year-over-year improvements in accuracy on image classification (from AlexNet\cite{AlexNet}, to ZFNet\cite{ZFNet}, to VGGNet\cite{VGGNet}, to ResNet\cite{ResNet}), which further led into improved accuracy on the other computer vision tasks. 
This also led to an upward trend in computation time as well as parameter count.
To mitigate this, starting in 2016 with SqueezeNet\cite{squeezenet}, Iandola et al. were successfully able to design networks that were 50 times smaller in parameters compared to AlexNet\cite{AlexNet}. MobileNets\cite{mobilenets} and ShuffleNet\cite{shufflenet} came soon after, optimizing their networks to have fewer computational operations, with the goal of reducing latency. The problem of reducing the size, the number of operations, and ultimately the latency of DNN inference became a widely-studied problem in computer vision research. One thing to note is that this research typically requires expertise in both computer vision as well as computer architecture.

\begin{figure*}
  \centering
  \includegraphics[width=0.85\textwidth]{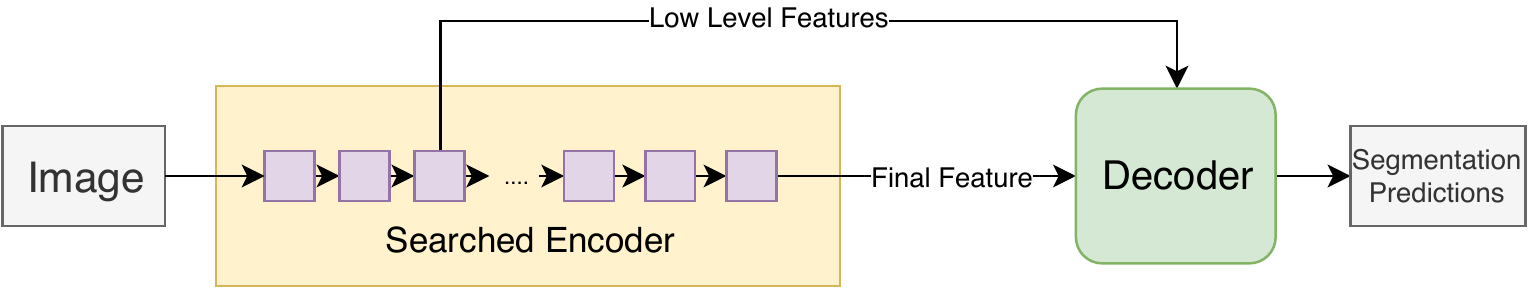}\\
    \centering
    \caption{General Encoder-Decoder Structure of our Segmentation Networks. We search the architecture space of the "Searched Encoder". We use either an ASPP\citep{deeplabv3} inspired decoder or the LR-ASPP Decoder depending on the search space.}
    \label{fig:architecture}
\vspace{-2mm}
\end{figure*}

\begin{figure}
    \vspace{-1mm}
    \centering
    \includegraphics[width=0.7\linewidth]{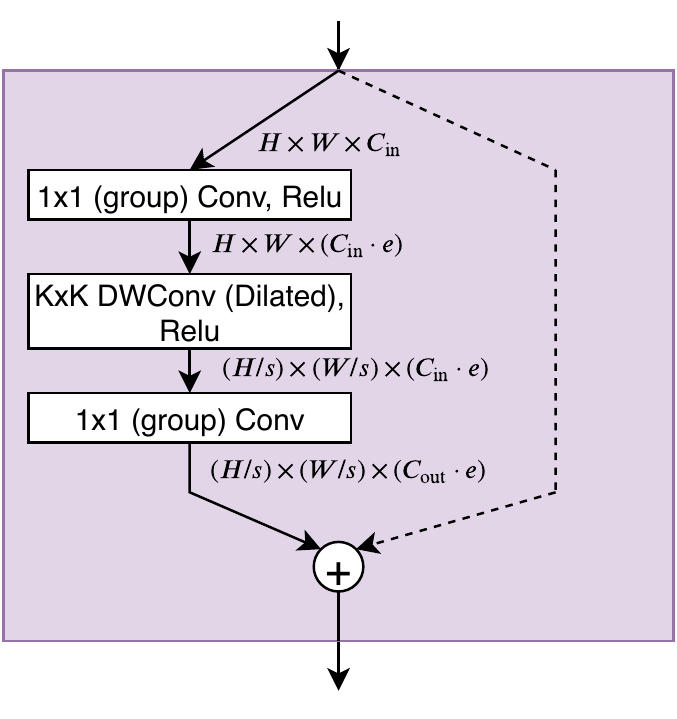}
    \vspace{-4mm}
    \caption{Diagram showing the architecture of the Inverted Residual blocks we use in our search space. They are parameterized so that the number of groups ($g\in\{1,2\}$) in the $1x1$ convolutions, the dilation rate of the depthwise convolution ($d \in \{1,2\}$), the kernel size ($k\in\{3,5\}$), and the expansion ratio ($e\in{1,3,6}$) may vary for different candidate blocks. The 12 possible configurations are shown in shown in Figure~\ref{fig:legend} and Appendix~\ref{appendix:candidate_blocks}. $C_\text{in}$, $C_\text{out}$, and stride ($s \in \{1,2\}$) are defined by the macro level parameters shown in Appendix~\ref{appendix:macro}. A residual connection is used if $C_\text{in} = C_\text{out}$ and $s=1$.}
    \label{fig:inverse_residual}
\vspace{-5mm}
\end{figure}

\subsection{Neural Architecture Search (NAS)}
\vspace{-0.08in} 
Since classification networks have commonly been used as the encoder for other computer vision tasks \cite{DPC, NASFPN, deeplabv3, RetinaNet, RCNN}, they are often a target of NAS searches\citep{DARTS, FBNET, SNAS, PROXYLESS, BASE, ENAS, PNAS, transferable} in efforts to exceed the performance of expert designed networks. However, many prior NAS works such as some that use Reinforcement Learning or Evolutionary search algorithms can often require thousands of GPU days per search\cite{transferable, MNAS, AmoebaNet}. 
The compute time of these searches would further increase if they were run directly on these high resolution vision tasks.
Howard et al. in MobileNetV3\cite{MobileNetV3} created networks for semantic segmentation by modifying classification networks that were produced by NAS. The NAS in that work had the objective of minimizing latency of the low resolution image classification network for mobile phones, and not for our ultimate goal of semantic segmentation at high resolution. 

Many works have developed methods to greatly reduce the search time of NAS\cite{EAS, ENAS, PNAS}. 
Recently, supernetwork-based NAS approaches have been proposed which have led to search times that are orders of magnitude faster by searching over millions of potential DNN designs while training just one supernetwork\cite{DARTS, FBNET, SNAS, PROXYLESS, BASE, autodeeplab}.
While there has been some work searching directly on other vision tasks, most of these do not also directly optimize for hardware latency\citep{NASFPN,autodeeplab, DPC}.
In our work described later in this paper, a gradient-based NAS method optimizes a supernetwork for both high semantic segmentation accuracy as well as low latency on our target hardware.
Our particular NAS algorithm utilizes the Gumbel-Softmax\cite{jang2017categorical} approximation of the categorical choice distribution which is also used in \cite{FBNET, SNAS, BASE}.

\section{Architecture Search Space}
\label{sec:architecture_search_space}
\vspace{-0.08in} 

In this work, we explore the space of encoders for semantic segmentation networks consisting of sequential Inverted Residual Blocks\citep{MobileNetV2}. The blocks are parameterized as shown in Figure~\ref{fig:inverse_residual}. In each architecture search, we constrain the macro-architecture and find optimal parameters for each block. This search space was chosen to be similar to the FBNet\cite{FBNET}, MobileNetV2\cite{MobileNetV2}, and MobileNetV3\citep{MobileNetV3} network families which allows us to directly compare our segmentation optimized networks to their classification optimized networks.

The general structure of all our networks is shown in Figure~\ref{fig:architecture}. We follow a common structure of some segmentation networks\citep{deeplabv3, MobileNetV3} where the decoder uses both the final output features from the encoder as well as a low level feature map from an earlier layer in the encoder. 

\subsection{Constrained Macro-Architecture}
\label{sec:fixed_macro}
\vspace{-0.08in} 

In our experiments we searched 3 search spaces: \texttt{Small}, \texttt{Large}, and \texttt{XLarge}.
To define each of these architecture spaces, we first constrain the macro-architecture of the encoder networks.
The macro-architectures describe the total number of blocks $N$ in the encoder, which decoder is used, and which layer our lower level features come from. For each block, we fix the input and output channels($C_{in}$ and $C_{out}$) and whether each block uses a stride of $s=1$ or $s=2$ in the depthwise convolution layer. It should be noted that since we allow each block to choose a no-op skip connection, the final layer count can be less than $N$.

The specifics of each of the three search spaces are shown in Appendix~\ref{appendix:macro}.
They were chosen to be comparable to the MobileNetV2\citep{MobileNetV2} and MobileNetV3\citep{MobileNetV3} segmentation networks.
In the \texttt{Small} and \texttt{Large} search spaces, we use the LR-ASPP\citep{MobileNetV3} decoder. In the \texttt{XLarge} search space, we use the variation of the ASPP decoder with fully depthwise convolutions proposed in \citet{DeepLabV3+}.

\begin{figure*}
  \includegraphics[width=0.75\textwidth]{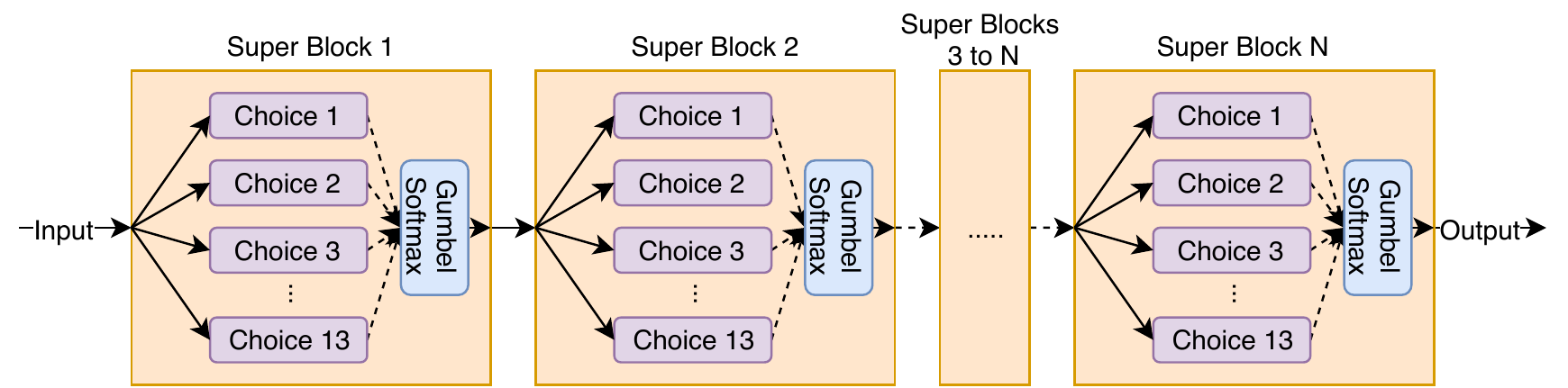}
    \centering
\vspace{-2mm}
    \caption{Diagram of a supernetwork with $N$ superblocks, which each contain 13 possible candidate block choices.}
    \label{fig:supernet}
\vspace{-2mm}
\end{figure*}

\begin{figure*}
  \includegraphics[width=0.6\textwidth]{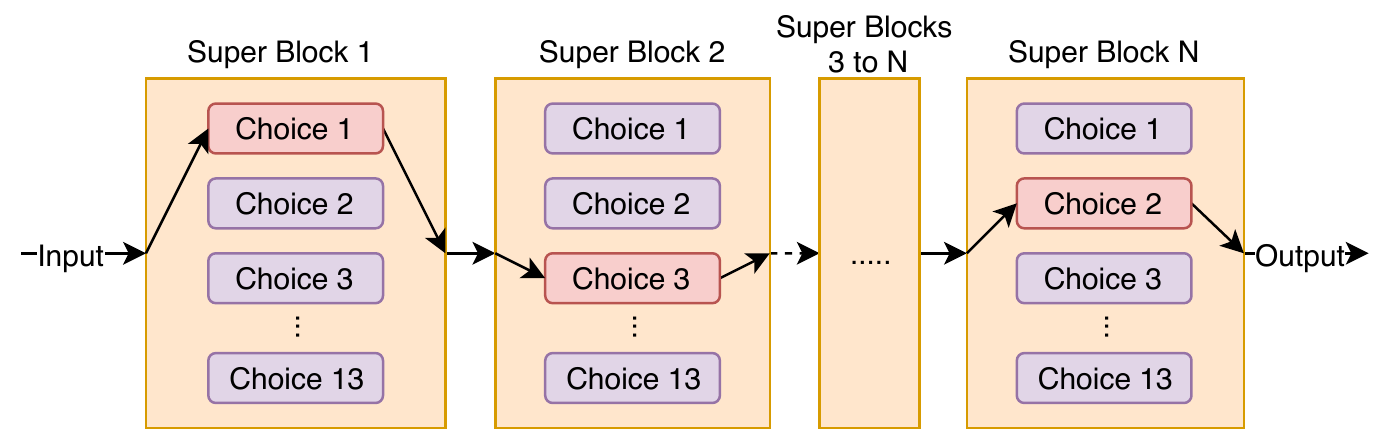}
    \centering
\vspace{-2mm}
    \caption{Diagram of an architecture path of a sampled architecture from a supernetwork. In this example, the 1st superblock uses candidate block 1, the 2nd superblock  uses candidate block 3, and the $N$th superblock uses candidate block 2.}
    \label{fig:supernet_inference}
\vspace{-2mm}
\end{figure*}

\subsection{Block Search Space}
\vspace{-0.08in} 

Within each macro-architecture space, our NAS picks the optimal hyperparameters for each block or replaces it with a no-op skip connection. As shown in Figure~\ref{fig:inverse_residual}, these hyperparameters define whether the 1x1 convolutions are grouped, whether the depthwise convolution is dilated with a rate 2, the size of the kernel $k$ for the depthwise convolution, and the expansion ratio $e$. We choose from 12 possible configurations as shown in Figure~\ref{fig:legend} and Appendix~\ref{appendix:candidate_blocks} as well as the skip connection.

\section{Neural Architecture Search Algorithm}
\label{sec:method}
\vspace{-0.08in} 
The particular approach and search space we use is similar to those used in \cite{FBNET}. We consider architecture search as a path-selection problem within a stochastic supernetwork such that any particular architecture in our search space is represented by some path through our supernetwork. As illustrated in Figure~\ref{fig:supernet}, we define our supernetwork to be a sequence of superblocks that each contain the candidate block choices. Running inference for a sampled architecture of the stochastic supernetwork is shown in Figure~\ref{fig:supernet_inference}.

We simultaneously co-optimize the convolutional weights ($w$) and architecture parameters ($\theta$) of the stochastic supernetwork to minimize our loss function which is defined as 
\begin{equation}
L(\theta,w) = L_P(\theta,w) + \alpha * L_E(\theta)
\end{equation}
where $L_P$ represents the problem-specific loss, $L_E$ is resource aware-loss term, and the hyperparameter $\alpha$ controls the tradeoff made between the two. As this work focuses on semantic segmentation, $L_P$ is a pixel-level cross-entropy loss. For $L_E$ we experiment with both the estimated total inference latency on our target-platform as well as the estimated number of Multiply-Accumulates for the network.

\subsection{Gumbel-Softmax}
\vspace{-0.08in} 

In order to make computation and optimization of the stochastic supernetwork tractable, each superblock picks a candidate block independent of the choices of other superblocks. Thus, we can model the choice of a candidate block as sampling from an independent categorical distribution where the probability of choosing candidate block $j$ for superblock $i$ in the network is $p(i,j)$. 
We define this probability using the softmax function on our architecture parameters ($\theta$) for each superblock.
\begin{equation}
    p(i,j | \theta) = \frac{e^{\theta_{i,j}}}{\sum_j^{13} e^{\theta_{i,j}}}
\end{equation}

The categorical distribution is difficult to directly optimize efficiently, so we use the Gumbel-Softmax relaxation of the categorical distribution proposed in \citet{jang2017categorical}. Sampling from the Gumbel-Softmax distribution allows us to efficiently optimize the architecture distribution by using gradient descent on the stochastic supernetwork. The Gumbel-Softmax distribution is controlled by a temperature parameter $t$. As $t$ approaches zero, the Gumbel-Softmax distribution becomes equivalent to the categorical distribution. The temperature parameter is annealed from 5.0 to 1.0 during our search.

\subsection{Early Stopping}
\vspace{-0.08in} 

A caveat of our supernetwork approach is that the optimization requires computation through every single candidate block for every iteration regardless of the learned architecture distribution. As optimal network architectures converge, the probability that a low performing candidate block is chosen decreases, but it still continues to use compute. So we use a compute optimization when the estimated probability of a candidate block being chosen is less than $0.5\%$. We simply remove it from the supernetwork. While there is some low probability that a removed candidate block could be optimal later in the search process, we have not seen this in practice. This compute optimization can cut search time in half.

\subsection{Resource-Aware Architecture Search}
\vspace{-0.08in} 

We define our resource aware loss as follows:
\begin{equation}
L_E(\theta) = \sum_{j}^N{ \sum_{i}^{13} p(i,j | \theta_i)C(i,j)}
\end{equation}

$C(i,j)$ represents the network resource cost of choosing candidate $j$ in block $i$ of the network. We model the resource cost of each block to be independent of others. $C$ can also be implemented as a lookup table similar to FBNet\cite{FBNET} so the resource costs only needs to be calculated once. Depending on how we build the lookup table, we can optimize for many different objectives ranging from hardware-agnostic metrics such as MACs or parameter size to hardware-aware costs like inference-time, memory accesses, or energy usage.

\begin{figure*}
    \begin{center}
    \begin{minipage}[t]{0.41\textwidth}
    \centering
        \centering
        \includegraphics[width=\textwidth]{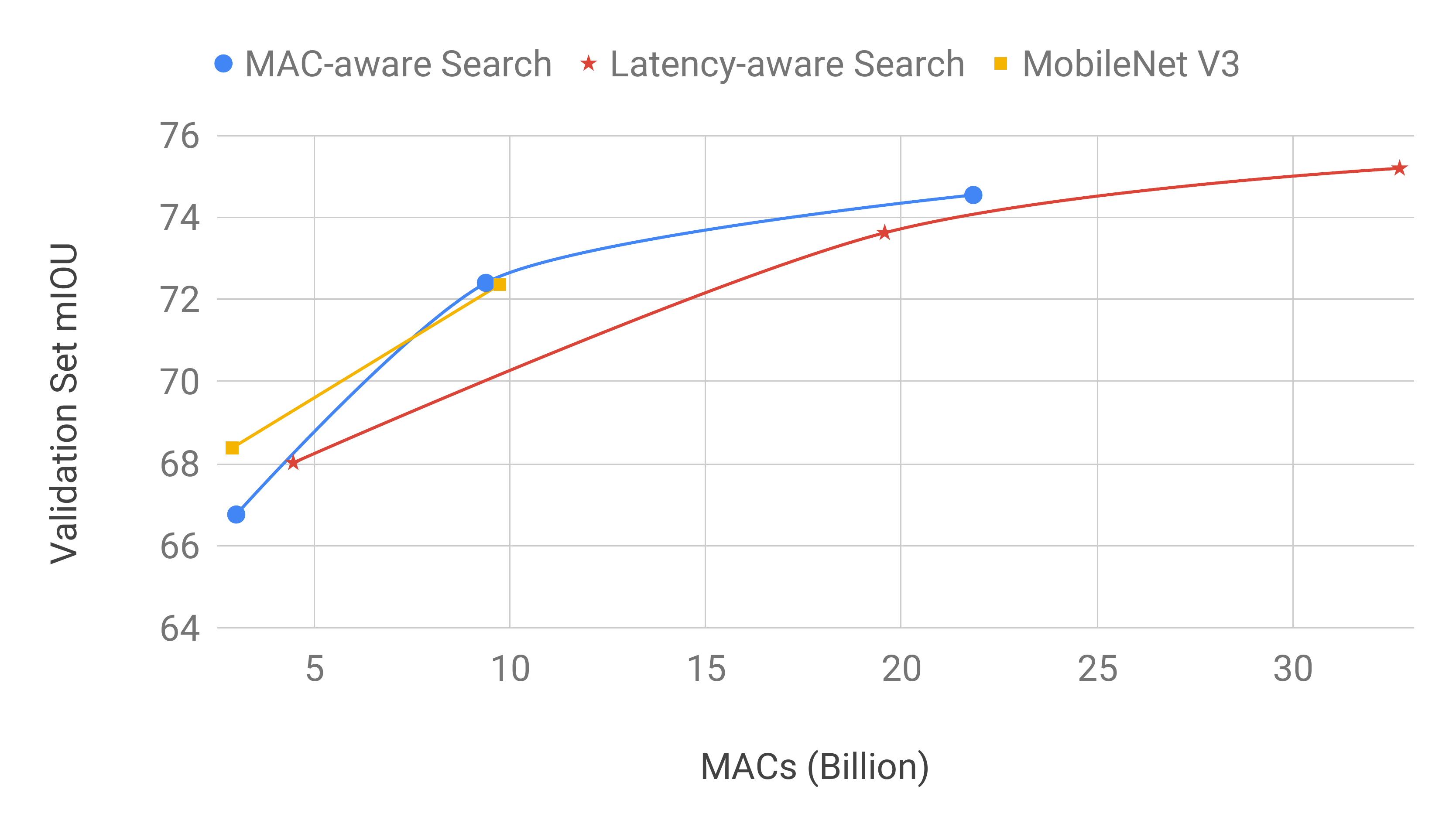}
        \vspace{-8mm}
        \caption{{\bf MACs} vs mIOU on Cityscapes validation set. SqueezeNAS MAC-optimized and latency-optimized models compared to MobileNetV3\cite{MobileNetV3} segmentation models.}
    \end{minipage}
    \hspace{10mm}
    \begin{minipage}[t]{0.41\textwidth}
        \centering
        \includegraphics[width=\textwidth]{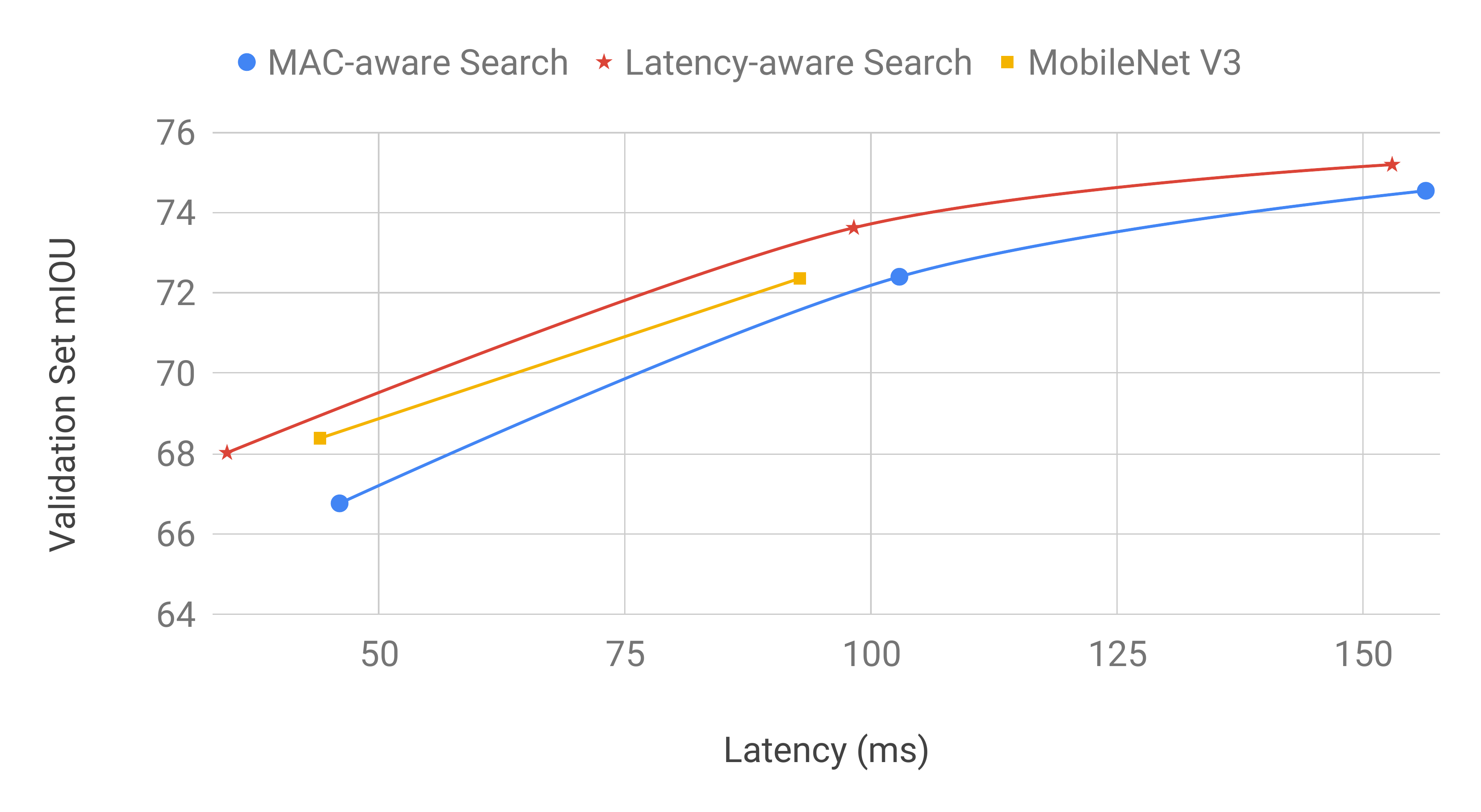}
        \vspace{-8mm}
        \caption{{\bf Latency} vs mIOU on Cityscapes validation set. SqueezeNAS MAC-optimized and latency-optimized models compared to MobileNetV3\cite{MobileNetV3} segmentation models.}
        \label{graph:lat_vs_iou}
    \end{minipage}
    \end{center}
\end{figure*}

\begin{table*}

  \vspace{-4mm}
  \centering
  \begin{tabular}{lllllll}
    \toprule        
    Architecture & Class mIOU & Latency (ms) & MACs (G) & MACs/sec (G) & Params (M) \\
    \midrule
    C3\cite{C3}                    & 61.96 & - & 6.29  & - & 0.19 \\
    EDANet\cite{EDANet}            & 65.11 & - & 8.97  & - & 0.68 \\
    MobileNetV2\cite{MobileNetV2} & 70.71 & - & 21.27 & - & 5.75 \\ 
    \midrule
    MobileNetV3-Small\cite{MobileNetV3} & 68.38 & 44.01 & 2.90 & 65.89 & 0.47 \\
    MobileNetV3-Large\cite{MobileNetV3} & 72.36 & 92.78 & 9.74 & 104.97 & 1.51 \\
    \midrule
    SqueezeNAS MAC Small & 66.76 & 46.01 & 3.01 & 65.37 & 0.30 \\
    SqueezeNAS MAC Large & 72.40 & 102.90 & 9.39 & 91.21 & 0.73 \\
    SqueezeNAS MAC XLarge & 74.54 & 156.41 & 21.84 & 139.63 & 1.80 \\
    \midrule
    SqueezeNAS LAT Small & 68.02 & 34.57 & 4.47 & 129.17 & 0.48 \\
    SqueezeNAS LAT Large & 73.62 & 98.28 & 19.57 & 199.17 & 1.90 \\
    SqueezeNAS LAT XLarge & 75.19 & 152.98 & 32.73 & 213.94 & 3.00 \\
    \bottomrule
  \end{tabular}
  \vspace{-2mm}
  \caption{Cityscapes Validation mIOU of MAC-Aware Searched, Latency-Aware Searched, and published state-of-the-art models. The latency values were benchmarked on the NVIDIA Xavier on the 30 watt power mode. Latency values for the MobileNetV3\cite{MobileNetV3} segmentation networks were obtained using an open source re-implementation.}
  \label{val-table}
    \vspace{-3mm}
\end{table*}

\begin{table*}
  \centering
  \begin{tabular}{llllll}
    \toprule
    Architecture & Class mIOU & Latency (ms) & MACs (Giga) & Params (M) \\
    \midrule
    MobileNet V3-Small\cite{MobileNetV3} & 69.4 & 44.01 & 2.90 & 0.47 \\
    MobileNet V3-Large\cite{MobileNetV3} & 72.6 & 92.78 & 9.74 & 1.51 \\
    \midrule
    SqueezeNAS LAT Small & 66.8 & 34.57 & 4.47 & 0.48 \\
    SqueezeNAS LAT Large & 72.5 & 98.28 & 19.57 & 1.90  \\
    \bottomrule
  \end{tabular}
    \vspace{-2mm}
  \caption{Test mIOU of Different Architectures on Cityscapes. The latency values were benchmarked on the NVIDIA Xavier on the 30 Watt power setting.}
  \label{test-table}
    \vspace{-3mm}
\end{table*}

\section{Experiments and Results}
\label{sec:experiments_and_results}
\vspace{-0.08in} 

We demonstrate two key ideas: first, Neural Architecture Search (NAS) is a powerful tool that can yield high-accuracy, low-latency networks. The second idea is that optimizing for hardware-agnostic metrics such as Multiply-Accumulates (MACs) is not an ideal proxy and can lead to sub-optimal latency results.

To demonstrate this, we use search spaces similar to prior work: the \texttt{Small}, \texttt{Large}, \texttt{XLarge} search spaces, which we define in Section \ref{sec:fixed_macro}. We first use our NAS method along with a hardware-agnostic objective (MACs) to generate a semantic segmentation network in each of our search spaces. These networks are comparable with current state-of-the-art networks on the MACs/Accuracy trade-off curve. We then measure the latency of these low-MAC networks on an embedded platform (NVIDIA Xaiver) as a baseline. Finally, we use our NAS method again on the same search spaces, but optimize with a hardware-aware objective (latency) to find 3 new networks targeted at similar latencies of the networks generated in the previous search. 

All search experiments are done on the Cityscapes\cite{Cityscapes} semantic segmentation dataset.

\subsection{Hardware-Agnostic Search}
\vspace{-0.08in} 

For our hardware-agnostic architecture searches, we apply our NAS method with a Multiply-Accumulates (MACs) minimization objective to create networks that are on the pareto-optimal tradeoff curve of MACs vs mIOU. To implement this, for each block $i$ in the network, we compute the number of Multiply-Accumulates for each candidate block $j$ and store the results in the lookup table $C$ such that $C(i,j) = MACS_{i,j}$.

We then perform an independent search in each of the three search spaces and obtain three MAC-optimized \textit{SqueezeNAS-MAC} networks. As shown in Table~\ref{val-table}, we achieve results that exceed the performance of prior work without NAS. 
We also achieve comparable results with MobileNetV3\cite{MobileNetV3} w.r.t the number of MACs. 
We finally measure the inference time of the 3 networks on a NVIDIA Xavier using cuDNN 7.3.1.
As in many applications requiring real-time inference, we use {\bf batch size = 1} for all of our latency tests throughout the paper. 
The results can be seen in Table~\ref{val-table}.

\subsection{Hardware-Aware Search}
\vspace{-0.08in} 

Our hardware-aware searches use the same NAS algorithm and architectural search space as the hardware-agnostic approach, but now we use a latency minimization objective for the resource-aware loss; formulated as $C(i,j) =  Latency_{i,j}$. To compute the latency of every candidate $j$ in each block $i$, we measure the inference time of all candidates on our target platform.
We conduct 3 new independent hardware-aware searches that target the latencies measured from the hardware-agnostic networks.
The results of these searches yield the three \textit{SqueezeNAS-LAT} networks. 
Our hardware-aware searches find networks that have significantly higher accuracies at the same or lower latency compared to the hardware-agnostic networks seen in Table~\ref{val-table}. The latency-optimized networks have a higher number of MACs, but they still run faster on our target device. 

\subsection{Implementation}
\vspace{-0.06in}

\subsubsection{Architecture Search}
\vspace{-0.08in} 
In our supernetwork-based architecture search, we train directly on the Cityscapes training set, without using any proxy task.
After we finish optimizing the supernetwork, we sample 200 discrete architectures from the optimal architecture distribution. We estimate the performance of each architecture by running inference on the Cityscapes fine validation dataset using the architecture path within the supernetwork as shown in Figure~\ref{fig:supernet_inference}. After validating the 200 architectures, we choose one from this estimated pareto-optimal frontier and retrain the singular architecture. The MAC-optimized networks are chosen to have comparable MACs to the MobileNetV3 segmentation networks, and the Latency-optimized networks are chosen to have inference latencies comparable with our MAC-optimized baseline networks.

\subsubsection{Training Details}
\label{sec:training_details}
\vspace{-0.08in} 

For comparability with other results, we follow a similar pretraining scheme to that used in \cite{DeepLabV3+}. 
After the architecture search is complete, we pretrain our sampled networks on ImageNet classification using the training regime used in ResNet\cite{ResNet}. We then do a stage of training on COCO~\cite{COCO} segmentation masks using the scheme used in DeepLabV3+\cite{DeepLabV3+}. 
Then, we train on the Cityscapes coarse training set annotations for 40 epochs, and finally we train on the Cityscapes fine training set annotations for 100 epochs, cutting the learning rate by 10 at 50 and 75 epochs. 
All segmentation training uses patch sizes of 768x768 pixels and are optimized with SGD with momentum, using a base learning rate of 0.05 and a weight-decay of 1e-5.

We use servers with 8 Nvidia Turing GPUs with 24GB of VRAM and train in mixed precision, allowing us to both leverage the tensor cores on the GPUs and fit a larger batch in VRAM. 
When we search larger supernetworks, we employ Synchronized BatchNorm\cite{Peng_2018_CVPR} to keep our BatchNorm\cite{batchnorm} batch sizes large enough for training stability.

\subsection{Results}
\vspace{-0.08in} 

First, our hardware-agnostic NAS method is able to produce networks that are competitive with the state-of-the-art with respect to both MACs and latency.
Compared to expert designed networks found without NAS such as EDANet~\cite{EDANet} and MobileNetV2~\cite{MobileNetV2}, our MAC-optimized networks achieve higher accuracy at a fraction of the MACs, as shown in Table~\ref{val-table}. 
Our \textit{SqueezeNAS-MAC-Small} network achieves more than 3\% higher absolute mIOU compared to the EDANet \cite{EDANet} segmentation network, which has three times more MACs than ours.
Our \textit{SqueezeNAS-MAC-Large} network achieves more than 2.5\% higher absolute mIOU compared to the MobileNetV2\cite{MobileNetV2} segmentation network, which has more than double the MACs of our network.
 
Our hardware-aware networks all have higher accuracy while having less latency compared to their hardware-agnostic counterparts. 
The \textit{SqueezeNAS-LAT-Small} network is 1.3\% more accurate, 35\% faster, and has 50\% more MACs compared to \textit{SqueezeNAS-MAC-Small}. 
The \textit{SqueezeNAS-LAT-Large} network is 1.2\% more accurate, 4\% faster, and has more than double the number of MACs compared to \textit{SqueezeNAS-MAC-Large}. 
This means that we're able to achieve double the number of operations in the same inference time window, as seen in Figure~\ref{graph:lat_vs_throughput}. 
This allows us to have much more expressive models that yield better accuracy while running at the same framerate.

We also compare our networks to the efficient segmentation networks proposed in MobileNetV3\citep{MobileNetV3}. These networks were optimized for image classification using NAS and were then modified for the semantic segmentation task.
The \textit{SqueezeNAS-MAC-Large} network is able to match the accuracy of the \textit{MobileNetV3-Large} network while using less MACs as seen in Table~\ref{val-table}.
It should be noted that the \textit{SqueezeNAS-MAC-Small} network does perform worse than MobileNetV3-Small. However, the MobileNetV3 networks do use Squeeze-Excitation\citep{hu2018squeeze} and Hard Swish\citep{swish} activations which our networks do not.
\textit{SqueezeNAS-LAT-Small} runs 20\% faster than \textit{MobileNetV3-Small} while achieving an mIOU that is only 0.26\% lower.
\textit{SqueezeNAS-LAT-Large} achieves over 1.2\% higher accuracy with less than 6\% higher latency.

We have noticed a small gap in our validation and test accuracies. This may be due to the small size of the Cityscapes dataset or the lack of our use of test-time augmentations.

The full validation set results are shown in Table~\ref{val-table}. Test set results are shown in Table~\ref{test-table}.
Each network was found in less than 15 GPU-days, which is more than 100 times less than some reinforcement learning and genetic search methods as shown in Table~\ref{tab:search-time}.

\begin{figure}
  \centering
    \centering
    \vspace{-2mm}
    \includegraphics[width=90mm]{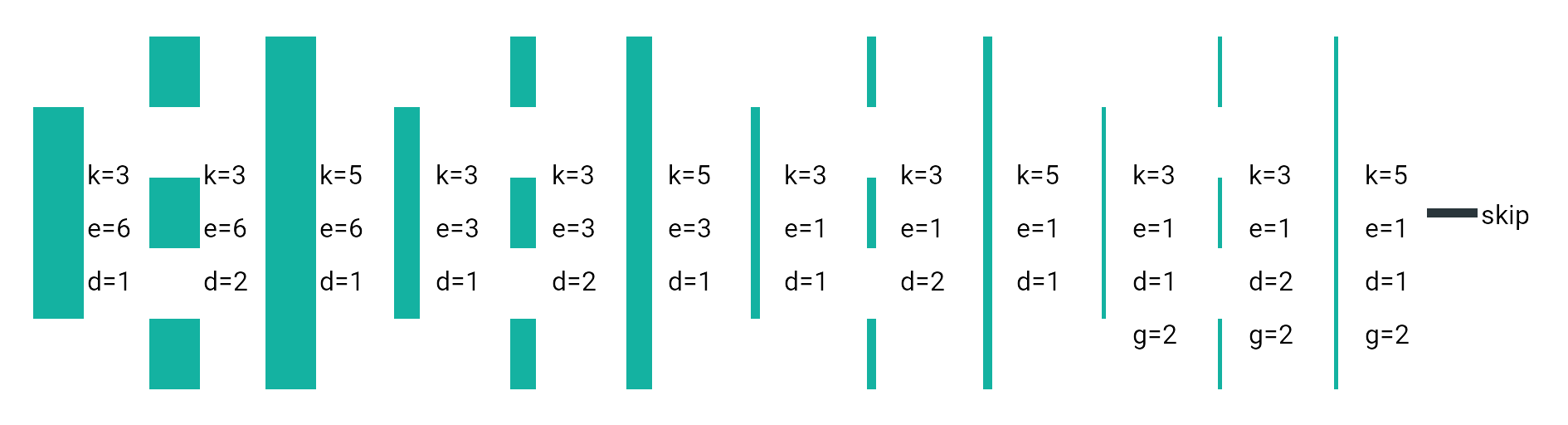}
    \vspace{-8mm}
    \caption{Visualization of the search space. Each of these blocks represent a MobileNetV2\cite{MobileNetV2} Inverted Residual block as seen in Figure~\ref{fig:inverse_residual}. \textit{k} represents the kernel size of the middle depthwise convolution layer. \textit{e} represents the expansion multiple for the depthwise convolution. \textit{d} represents the dilation rate of the depthwise convolution. \textit{g} represents the number of groups(1 if not listed) in the 1x1 convolutions. Finally we have a no-op \textit{skip} connection that can be chosen.}
    \label{fig:legend}
    \centering
    \vspace{3mm}
    \includegraphics[width=90mm]{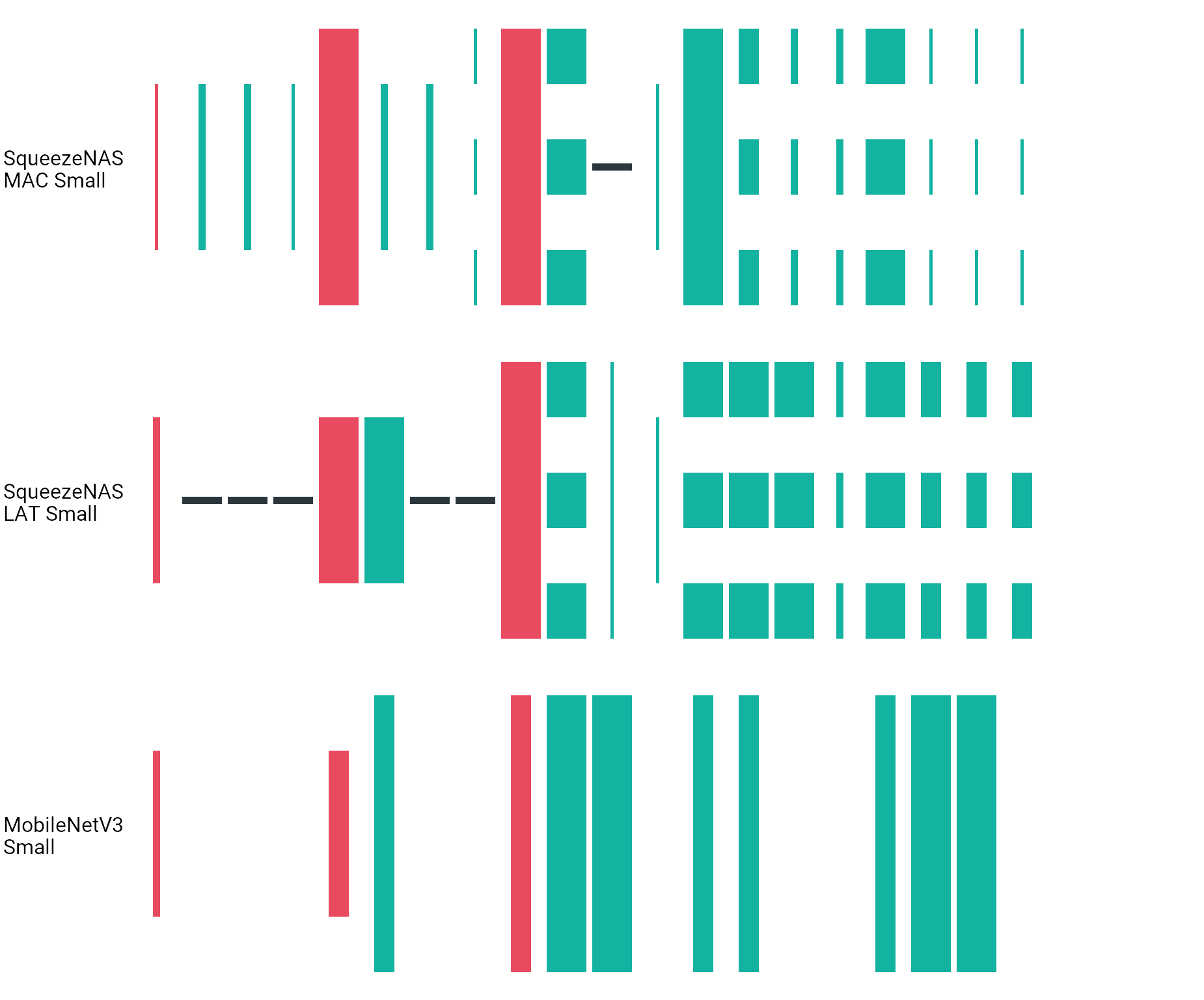}
    \vspace{-8mm}
    \caption{Small Networks. Networks are lined up at their down-sampling block represented by the color red.}
    \label{fig:small}
\vspace{-3mm}
\end{figure}  
  
\begin{figure}[h]
    \centering
    \includegraphics[width=90mm]{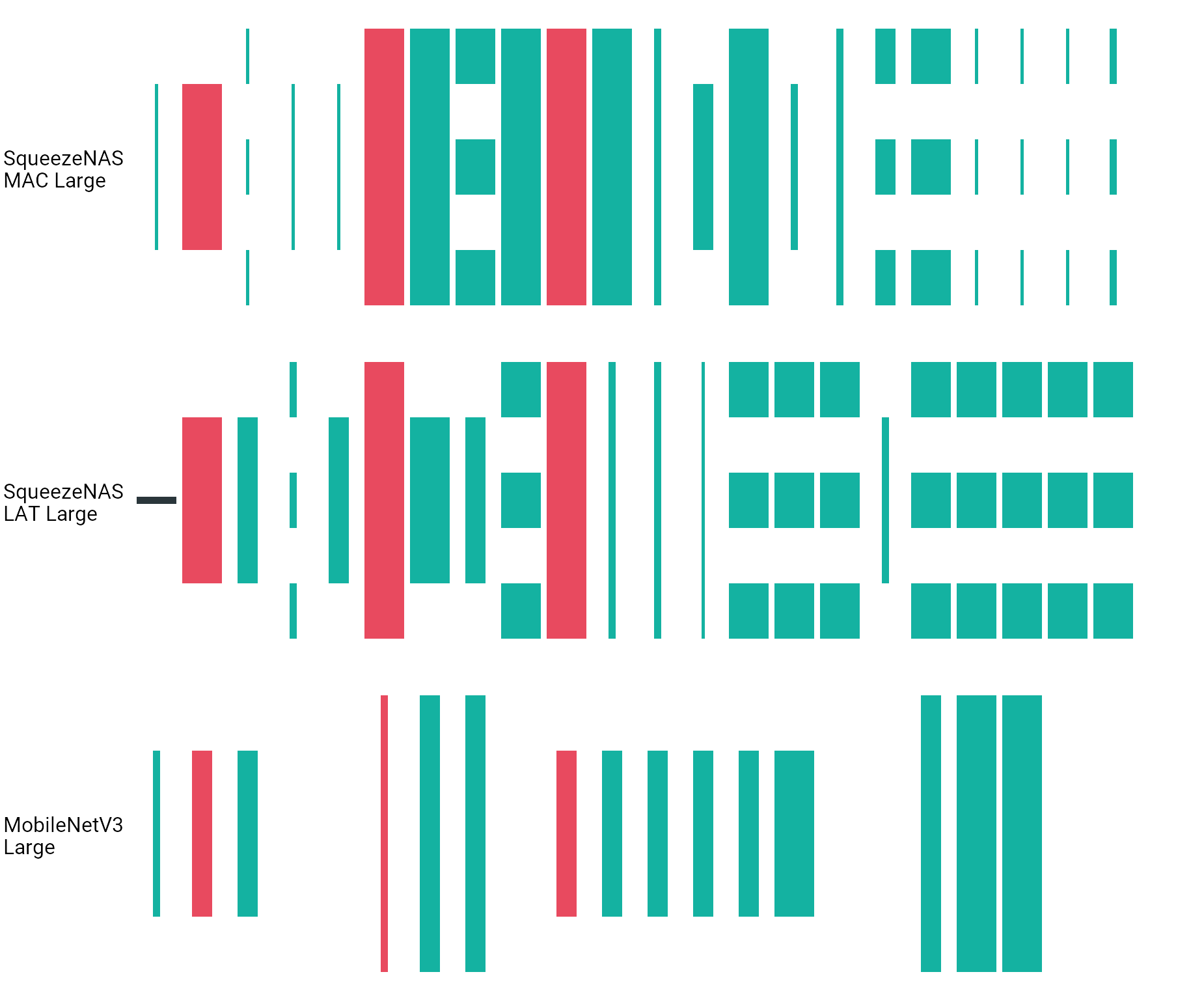}
    \vspace{-8mm}
    \caption{Large Networks. Networks are lined up at their down-sampling block represented by the color red.}
    \label{fig:large}
    \vspace{-4mm}
\end{figure}

\begin{savenotes}
\begin{table}
    \centering
    \begin{tabular}{lll}
        \toprule
        Architecture & Search Time (GPU Days) \\
        \midrule
        NAS with RL\cite{zoph2016neural} & 22,400 \\
        NASNet\cite{transferable} & 2,000 \\
        MnasNet\cite{MNAS} & 2,000 \footnote{Approximated from TPUv2 Hours} \\
        MobileNetV3\cite{MobileNetV3} & $>$ 2,000 \footnote{Starts with a MnasNet network  (search time is approximated from TPUv2 Hours) and adapts it with the NetAdapt NAS algorithm. The NetAdapt search time is not included since it is not reported in the paper\cite{MobileNetV3}.} \\ 
        AmoebaNet\cite{AmoebaNet} & 3,150 \\
        FBNet\cite{FBNET} & 9 \\
        DARTS\cite{DARTS} & 4 \\
        \midrule
        SqueezeNAS MAC Small & 7.0 \\
        SqueezeNAS MAC Large & 9.7 \\
        SqueezeNAS MAC XLarge & 14.6 \\
        \midrule
        SqueezeNAS LAT Small & 8.7 \\
        SqueezeNAS LAT Large & 9.4 \\
        SqueezeNAS LAT XLarge & 11.5 \\
        \bottomrule
    
    \end{tabular}
    \vspace{-2mm}
    \caption{Search times of SqueezeNAS Networks compared to other NAS methods.}
    \label{tab:search-time}
    \vspace{-4mm}
\end{table}
\end{savenotes}
\section{Network Analysis}
\label{sec:network_analysis}
\vspace{-0.08in} 

We now compare the block choices of the hardware-agnostic, hardware-aware, and MobileNetV3 segmentation networks. Since the three families all use the same Inverted Residual blocks, we can place MobileNetV3's building blocks into our 13 candidate blocks which can be seen in Figure~\ref{fig:legend}. One caveat to note is that we are not accounting for the Squeeze-Excitation\cite{hu2018squeeze} blocks that are in some MobileNetV3 blocks for visualization, and the expansion ratios are approximated to be either 1, 3, or 6.

We visualize the small networks in Figure~\ref{fig:small}. We first examine our \textit{SqueezeNAS-MAC-Small} network and see that it uses a mix of low and high expansion blocks. It also uses the highest compute candidate block possible for its second and third downsampling blocks. The last thing to note is that our NAS method chose to use dilated \textit{3x3} blocks for the last stage of the network. This is a very common trend that we see in expert designed, high resolution semantic segmentation networks such as DeepLabV3\cite{deeplabv3} and PSPNet\cite{PSPNet}.

The next small network we examine is our \textit{SqueezeNAS-LAT-Small}, which is more accurate and lower latency than the previous network. A radical difference that we immediately see is that the network uses many more skip connections instead of low expansion blocks. This makes the macro-architecture look very similar to that of \textit{MobileNetV3-Small}, also visualized in Figure~\ref{fig:small}. Both networks do aggressive down-sampling and push their compute (via higher expansion ratios) later in the network, where the resolution is lower and the base channel count is higher. This yields a higher arithmetic-intensity.\footnote{Arithmetic Intensity is the ratio of MACs to memory traffic~\cite{roofline}. When arithmetic intensity drops below a certain threshold, the latency is dominated by the time to access data from memory.} On devices like GPUs, which are typically memory bandwidth bound, higher arithmetic-intensity allows for more operations for the same memory bandwidth. It is interesting to see how both of the latency optimizing NAS methods produce similar networks that follow intuition from a computer architecture perspective. The networks differ in that our network uses more blocks but with a smaller kernel sizes near the end of the network. (\textit{3x3} dilated vs \textit{5x5}). Which is consistent with our hardware-agnostic network and other related segmentation work.

\begin{figure}
 \centering
 \includegraphics[width=80mm]{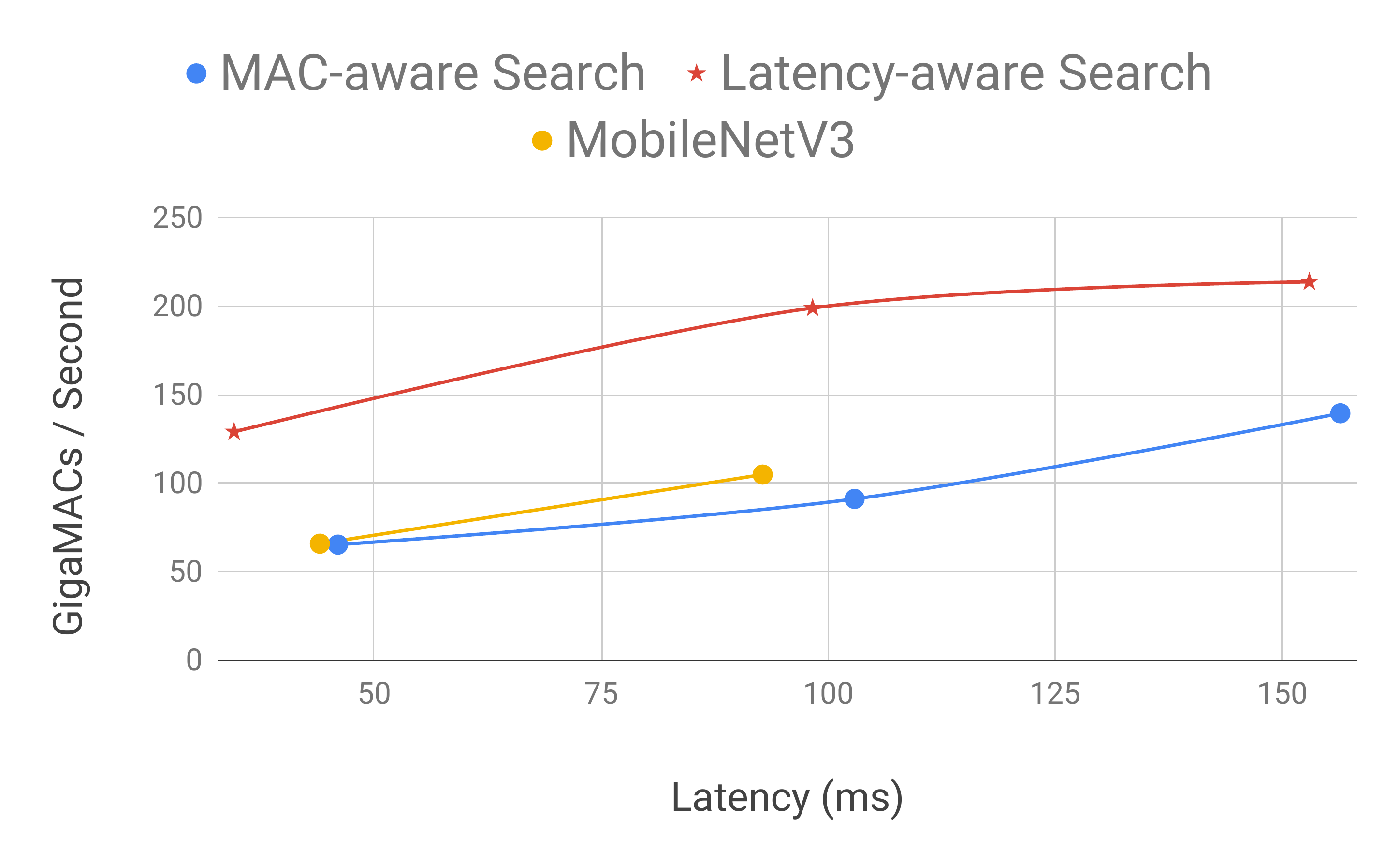}
 \vspace{-5mm}
 \caption{Comparison of Throughput (GigaMACs per second) vs Latency, of SqueezeNAS networks and MobileNetV3\cite{MobileNetV3} segmentation networks.}
 \label{graph:lat_vs_throughput}
 \vspace{-3mm}
\end{figure}

We now visually compare the large networks in Figure~\ref{fig:large}. Both \textit{SqueezeNAS-MAC-Large} and \textit{SqueezeNAS-LAT-Large}, follow the a trend similar to our smaller networks where they all have high compute down-sampling blocks, as well as heavy use of dilated convolutions in the second half of the networks. If we compare the MAC and latency networks, we see that the MAC network has the majority of its compute in the middle, whereas the latency network pushes its compute towards the end where it would yield a higher overall arithmetic-intensity for the network. This also has the side-effect of more than doubling the total number of MACs but still decreasing latency. We can conclude with saying that our NAS method is effective at producing high-throughput networks while maintaining low latency as seen in Figure~\ref{graph:lat_vs_throughput}.

\section{Conclusion}
\label{sec:conclusion}
\vspace{-0.08in} 

In Section~\ref{sec:introduction_and_motivation}, we presented a playbook for replacing architecture-transfer with neural architecture search to develop DNNs that are optimized for specific tasks and for specific computing platforms.
After following this playbook throughout this paper, we have learned the following.

First, by doing a proxyless search on a semantic segmentation dataset, our NAS produced the {\em SqueezeNAS} family of models, which achieve superior latency-accuracy tradeoffs relative to MobileNetV3 on the semantic segmentation validation set. 
We attribute our superior results, at least in part, to the fact that the backbone of the MobileNetV3 semantic segmentation network was designed by NAS for the proxy task of image classification on mobile phones (that is to say, it was not designed in a proxyless manner for semantic segmentation on embedded GPU devices).

Second, while the MobileNetV3 authors searched for thousands of GPU days, our approach produced these results in 7 to 15 GPU days per search. 
In other words, modern supernetwork-based NAS can now produce state-of-the-art results in less than a weekend of search time on an 8 GPU server.

Third, recall that we did two sets of NAS experiments: one in which we searched for low-MAC models, and one where we searched for low-latency models on a target computing platform. 
We achieved substantially faster and more accurate models when searching for latency on the target platform. 
Finally, given the growing diversity of chips and computing platforms designed for deep neural networks, we believe that using NAS to optimize for low latency on a target computing platform will continue to grow in importance. 

\clearpage
\bibliography{ms}

\clearpage
\newpage

\appendix
\onecolumn

\begin{appendix}

\thispagestyle{plain}
\begin{center}
{\Large \bf Appendix}
\end{center}

\section{XLarge Networks Visualization}
\label{appendix:xlarge_networks}
\vspace{-0.1in}

\begin{center}
\includegraphics[width=95mm]{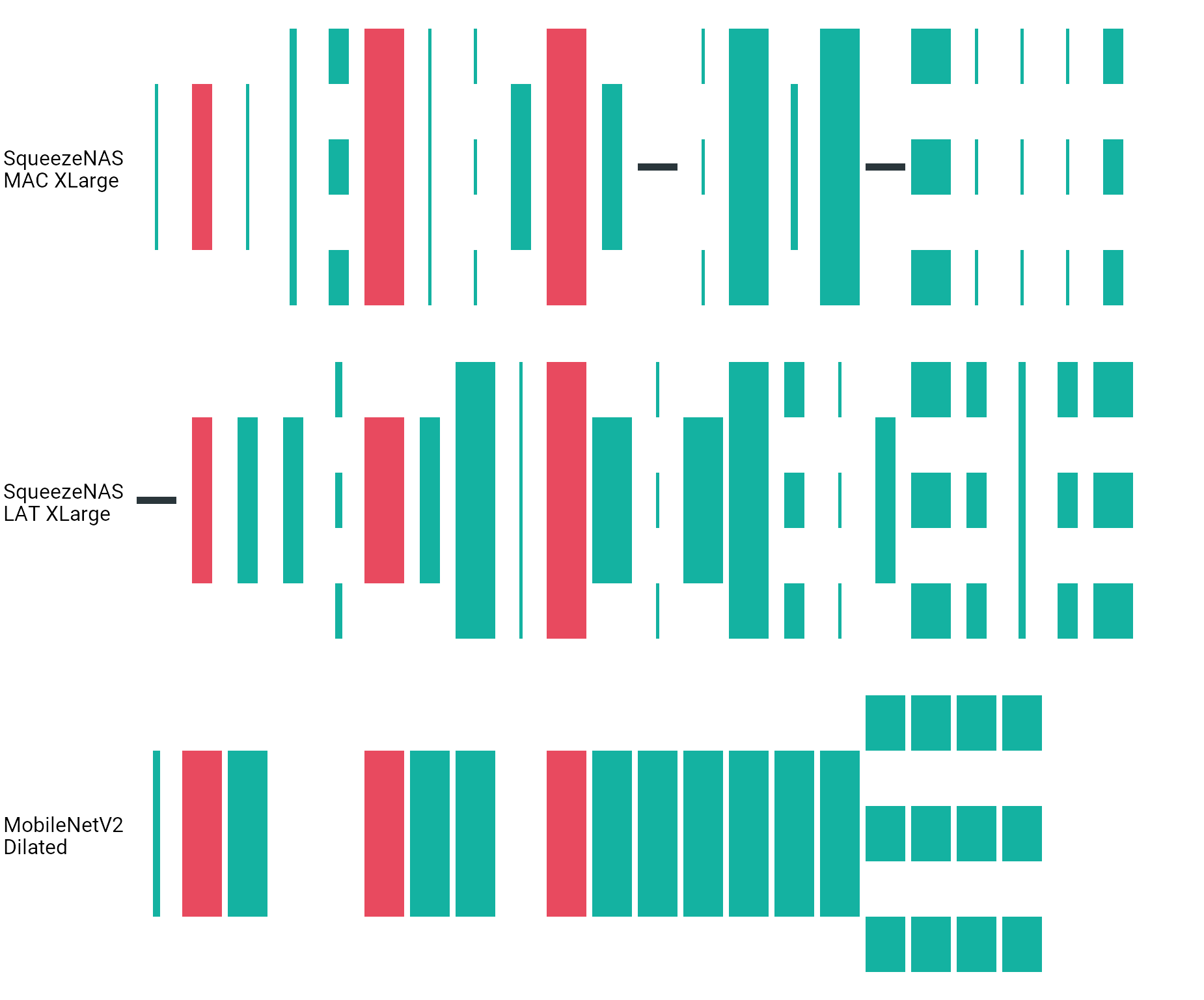}\\

\end{center}
XLarge Networks. Networks are lined up at their down-sampling block represented by the color red. We show the MobileNetV2 segmentation encoder with an output stride of 16 for comparison.
\\\\\\

\section{Parameters of Candidate Blocks in our Search Space}\label{appendix:candidate_blocks}
\begin{center}
  \centering
  \begin{tabular}{ccccc}
    \toprule
    Block type & \thead{Kernel\\($K$)} & Dilation & \thead{Expand\\($e$)} & Group \\ 
    \midrule
    k3\_d1\_e1\_g2 & 3 & 1 & 1 & 2 \\
    k3\_d1\_e1\_g1 & 3 & 1 & 1 & 1 \\
    k3\_d1\_e3\_g1 & 3 & 1 & 3 & 1 \\
    k3\_d1\_e6\_g1 & 3 & 1 & 6 & 1 \\
    k3\_d2\_e1\_g2 & 3 & 2 & 1 & 2 \\
    k3\_d2\_e1\_g1 & 3 & 2 & 1 & 1 \\
    k3\_d2\_e3\_g1 & 3 & 2 & 3 & 1 \\
    k3\_d2\_e6\_g1 & 3 & 2 & 6 & 1 \\
    k5\_d1\_e1\_g2 & 5 & 1 & 1 & 2 \\
    k5\_d1\_e1\_g1 & 5 & 1 & 1 & 1 \\
    k5\_d1\_e3\_g1 & 5 & 1 & 3 & 1 \\
    k5\_d1\_e6\_g1 & 5 & 1 & 6 & 1 \\
    skip & - & - & - & -\\
    \bottomrule
  \end{tabular}
\end{center}

\clearpage
\section{Parameters of Macro Search Spaces}\label{appendix:macro}
\subsection{Small Network Search Space}
\begin{center}
Encoder Macro Architecture and found MAC/Latency optimized networks\\
  \begin{tabular}{ccccccc}
    \toprule
    Operator & $C_{in}$ & $C_{out}$ & $s$ & Output Stride & MAC Network Layer & Latency Network Layer\\ 
    \midrule
    conv2d, 3x3 & 3 & 16 & 2 & 2 & - & - \\
    Searched Inverted Residual & 16 & 16 & 2 & 4 & k3\_d1\_e1\_g2 &  k3\_d1\_e1\_g1\\
    Searched Inverted Residual & 16 & 16 & 1 & 4 & k3\_d1\_e1\_g1 & skip\\
    Searched Inverted Residual & 16 & 16 & 1 & 4 & k3\_d1\_e1\_g1 & skip\\
    Searched Inverted Residual & 16 & 16 & 1 & 4 & k3\_d1\_e1\_g2 & skip\\
    Searched Inverted Residual & 16 & 24 & 2 & 8 & k5\_d1\_e6\_g1 & k3\_d1\_e6\_g1\\
    Searched Inverted Residual & 24 & 24 & 1 & 8 & k3\_d1\_e1\_g1 & k3\_d1\_e6\_g1\\
    Searched Inverted Residual & 24 & 24 & 1 & 8 & k3\_d1\_e1\_g1 & skip\\
    Searched Inverted Residual & 24 & 24 & 1 & 8 & k3\_d2\_e1\_g2 & skip\\
    Searched Inverted Residual & 24 & 40 & 2 & 16 & k5\_d1\_e6\_g1 & k5\_d1\_e6\_g1\\
    Searched Inverted Residual & 40 & 40 & 1 & 16 & k3\_d2\_e6\_g1 & k3\_d2\_e6\_g1\\
    Searched Inverted Residual & 40 & 40 & 1 & 16 & skip & k5\_d1\_e1\_g2\\
    Searched Inverted Residual & 40 & 40 & 1 & 16 & k3\_d1\_e1\_g2 & k3\_d1\_e1\_g2\\
    Searched Inverted Residual & 40 & 48 & 1 & 16 & k5\_d1\_e6\_g1 & k3\_d2\_e6\_g1\\
    Searched Inverted Residual & 48 & 48 & 1 & 16 & k3\_d2\_e3\_g1 & k3\_d2\_e6\_g1\\
    Searched Inverted Residual & 48 & 48 & 1 & 16 & k3\_d2\_e1\_g1 & k3\_d2\_e6\_g1 \\
    Searched Inverted Residual & 48 & 48 & 1 & 16 & k3\_d2\_e1\_g1 & k3\_d2\_e1\_g1\\
    Searched Inverted Residual & 48 & 96 & 1 & 16 & k3\_d2\_e6\_g1 & k3\_d2\_e6\_g1\\
    Searched Inverted Residual & 96 & 96 & 1 & 16 & k3\_d2\_e1\_g2 & k3\_d2\_e3\_g1\\
    Searched Inverted Residual & 96 & 96 & 1 & 16 & k3\_d2\_e1\_g2 & k3\_d2\_e3\_g1\\
    Searched Inverted Residual & 96 & 96 & 1 & 16 & k3\_d2\_e1\_g2 & k3\_d2\_e3\_g1\\
    \bottomrule
  \end{tabular}\\
  
\end{center}
The decoder used was the LR-ASPP proposed in \citet{MobileNetV3} with 128 channels in its layers. 
It uses the lower level feature from the last block with output stride 8. 
The "MAC Network Layer" and  "Latency Network Layer" indicate the layers chosen by our MAC and Latency optimized Neural Architecture Searches Respectively.

\vspace{6mm}
\subsection{Large Network Search Space}
\vspace{-0.1in}
\begin{center}
Encoder Macro Architecture and found MAC/Latency optimized networks\\
  \begin{tabular}{ccccccc}
    \toprule
    Operator & $C_{in}$ & $C_{out}$ & $s$ & Output Stride & MAC Network Layer & Latency Network Layer \\ 
    \midrule
    conv2d, 3x3 & 3 & 16 & 2 & 2 & - & - \\
    Searched Inverted Residual & 16 & 24 & 1 & 4 & k3\_d1\_e1\_g2 & skip \\
    Searched Inverted Residual & 24 & 32 & 2 & 4 & k3\_d1\_e6\_g1 & k3\_d1\_e6\_g1\\
    Searched Inverted Residual & 32 & 32 & 1 & 4 & k3\_d2\_e1\_g2 & k3\_d1\_e3\_g1\\
    Searched Inverted Residual & 32 & 32 & 1 & 4 & k3\_d1\_e1\_g2 & k3\_d2\_e1\_g1\\
    Searched Inverted Residual & 32 & 32 & 1 & 4 & k3\_d1\_e1\_g2 & k3\_d1\_e3\_g1\\
    Searched Inverted Residual & 32 & 48 & 2 & 8 & k5\_d1\_e6\_g1 & k5\_d1\_e6\_g1\\
    Searched Inverted Residual & 48 & 48 & 1 & 8 & k5\_d1\_e6\_g1 & k3\_d1\_e6\_g1\\
    Searched Inverted Residual & 48 & 48 & 1 & 8 & k3\_d2\_e6\_g1 & k3\_d1\_e3\_g1\\
    Searched Inverted Residual & 48 & 48 & 1 & 8 & k5\_d1\_e6\_g1 & k3\_d2\_e6\_g1\\
    Searched Inverted Residual & 48 & 96 & 2 & 16 & k5\_d1\_e6\_g1 & k5\_d1\_e6\_g1\\
    Searched Inverted Residual & 96 & 96 & 1 & 16 & k5\_d1\_e6\_g1 & k5\_d1\_e1\_g1\\
    Searched Inverted Residual & 96 & 96 & 1 & 16 & k5\_d1\_e1\_g1 & k5\_d1\_e1\_g1\\
    Searched Inverted Residual & 96 & 96 & 1 & 16 & k3\_d1\_e3\_g1 & k5\_d1\_e1\_g2\\
    Searched Inverted Residual & 96 & 144 & 1 & 16 & k5\_d1\_e6\_g1 & k3\_d2\_e6\_g1\\
    Searched Inverted Residual & 144 & 144 & 1 & 16 & k3\_d1\_e1\_g1 & k3\_d2\_e6\_g1\\
    Searched Inverted Residual & 144 & 144 & 1 & 16 & k5\_d1\_e1\_g1 & k3\_d2\_e6\_g1\\
    Searched Inverted Residual & 144 & 144 & 1 & 16 & k3\_d2\_e3\_g1 & k3\_d1\_e1\_g1\\
    Searched Inverted Residual & 144 & 240 & 1 & 16 & k3\_d2\_e6\_g1 & k3\_d2\_e6\_g1\\
    Searched Inverted Residual & 240 & 240 & 1 & 16 & k3\_d2\_e1\_g2 & k3\_d2\_e6\_g1\\
    Searched Inverted Residual & 240 & 240 & 1 & 16 & k3\_d2\_e1\_g2 & k3\_d2\_e6\_g1\\
    Searched Inverted Residual & 240 & 240 & 1 & 16 & k3\_d2\_e1\_g2 & k3\_d2\_e6\_g1\\
    Searched Inverted Residual & 240 & 240 & 1 & 16 & k3\_d2\_e1\_g1 & k3\_d2\_e6\_g1\\
    \bottomrule
  \end{tabular}\\
  
\end{center}
The decoder used was the LR-ASPP proposed in \citet{MobileNetV3} with 128 channels in its layers.
For the lower level feature we used the output of the last block with output stride 8. 
The "MAC Network Layer" and  "Latency Network Layer" indicate the layers chosen by our MAC and Latency optimized Neural Architecture Searches Respectively.

\clearpage
\subsection{XLarge Network Search Space}
\vspace{-0.1in}
\begin{center}
Encoder Macro Architecture and found MAC/Latency optimized networks\\
  \begin{tabular}{ccccccc}
    \toprule
    Operator & $C_{in}$ & $C_{out}$ & $s$ & Output Stride  & MAC Network Layer & Latency Network Layer \\ 
    \midrule
    conv2d, 3x3 & 3 & 16 & 2 & 2 & - & - \\
    Searched Inverted Residual & 16 & 24 & 1 & 4 & k3\_d1\_e1\_g2 & skip\\
    Searched Inverted Residual & 24 & 24 & 2 & 4 & k3\_d1\_e3\_g1 & k3\_d1\_e3\_g1\\
    Searched Inverted Residual & 24 & 24 & 1 & 4 & k3\_d1\_e1\_g2 & k3\_d1\_e3\_g1\\
    Searched Inverted Residual & 24 & 24 & 1 & 4 & k5\_d1\_e1\_g1 & k3\_d1\_e3\_g1\\
    Searched Inverted Residual & 24 & 24 & 1 & 4 & k3\_d2\_e3\_g1 & k3\_d2\_e1\_g1\\
    Searched Inverted Residual & 24 & 32 & 2 & 8 & k5\_d1\_e6\_g1 & k3\_d1\_e6\_g1\\
    Searched Inverted Residual & 32 & 32 & 1 & 8 & k5\_d1\_e1\_g2 & k3\_d1\_e3\_g1\\
    Searched Inverted Residual & 32 & 32 & 1 & 8 & k3\_d2\_e1\_g2 & k5\_d1\_e6\_g1\\
    Searched Inverted Residual & 32 & 32 & 1 & 8 & k3\_d1\_e3\_g1 & k5\_d1\_e1\_g2\\
    Searched Inverted Residual & 32 & 64 & 2 & 16 & k5\_d1\_e6\_g1 & k5\_d1\_e6\_g1\\
    Searched Inverted Residual & 64 & 64 & 1 & 16 & k3\_d1\_e3\_g1 & k3\_d1\_e6\_g1\\
    Searched Inverted Residual & 64 & 64 & 1 & 16 & skip & k3\_d2\_e1\_g2\\
    Searched Inverted Residual & 64 & 64 & 1 & 16 & k3\_d2\_e1\_g2 & k3\_d1\_e6\_g1\\
    Searched Inverted Residual & 64 & 96 & 1 & 16 & k5\_d1\_e6\_g1 & k5\_d1\_e6\_g1\\
    Searched Inverted Residual & 96 & 96 & 1 & 16 & k3\_d1\_e1\_g1 & k3\_d2\_e3\_g1\\
    Searched Inverted Residual & 96 & 96 & 1 & 16 & k5\_d1\_e6\_g1 & k3\_d2\_e1\_g2\\
    Searched Inverted Residual & 96 & 96 & 1 & 16 & skip & k3\_d1\_e3\_g1\\
    Searched Inverted Residual & 96 & 160 & 1 & 16 & k3\_d2\_e6\_g1 & k3\_d2\_e6\_g1\\
    Searched Inverted Residual & 160 & 160 & 1 & 16 & k3\_d2\_e1\_g2 & k3\_d2\_e3\_g1\\
    Searched Inverted Residual & 160 & 160 & 1 & 16 & k3\_d2\_e1\_g2 & k5\_d1\_e1\_g1\\
    Searched Inverted Residual & 160 & 160 & 1 & 16 & k3\_d2\_e1\_g2 & k3\_d2\_e3\_g1\\
    Searched Inverted Residual & 160 & 160 & 1 & 16 & k3\_d2\_e3\_g1 & k3\_d2\_e6\_g1\\
    conv2d, 1x1 & 96 & 256 & 1 & 16 & - & -\\
    \bottomrule
  \end{tabular}\\
\end{center}
The decoder used is the ASPP with fully depthwise convolutions proposed in \citet{deeplabv3}.
It uses the lower level feature from the last block with output stride 4.
The "MAC Network Layer" and  "Latency Network Layer" indicate the layers chosen by our MAC and Latency optimized Neural Architecture Searches Respectively.

\end{appendix}
\end{document}